\documentclass[journal]{IEEEtran}
\usepackage{microtype} 
\usepackage{multirow}
\usepackage{graphicx}
\usepackage{booktabs}
\usepackage{amsmath}
\usepackage{mathtools}
\usepackage{hyperref}
\usepackage{balance}
\usepackage{amsfonts}       
\usepackage{nicefrac}       
\usepackage{xcolor}  
\usepackage{soul}
\usepackage{kotex}     
\usepackage{wrapfig}
\usepackage{enumitem}
\usepackage{float}
\usepackage{physics}
\usepackage{subfigure}
\usepackage{bm}
\usepackage{bbm}
\usepackage{textcomp}
\usepackage{tikz}
\usepackage{amsthm}
\usepackage{thmtools}
\usepackage{thm-restate}
\usepackage{amssymb}
\usepackage{qcircuit}
\usepackage{array}
\usepackage[algo2e, linesnumbered, ruled, vlined]{algorithm2e}
\usepackage{etoolbox}

\newtheorem{theorem}{Theorem}

\declaretheorem[name=Lemma]{lemma}

\newcommand{\thetab}{\ensuremath{\bm{\theta}}}

\newcommand{\BfPara}[1]{{\noindent\bf#1.}\xspace}

\usepackage[normalem]{ulem}

\begin{document}
\title{Fast Quantum Convolutional Neural Networks for Low-Complexity Object Detection in Autonomous Driving Applications} %
\author{
    Hankyul Baek, Donghyeon Kim, and
    Joongheon Kim,~\IEEEmembership{Senior Member,~IEEE}
    \thanks{This research was funded by Institute of Advanced Technology Development (IATD) in Hyundai Motor Company and Kia Corporation. \textit{(Corresponding authors: Donghyeon Kim, Joongheon Kim)}}
    \thanks{H. Baek and J. Kim are with the Department of Electrical and Computer Engineering, Korea University, Seoul 02841, Republic of Korea (e-mails: \{67back, joongheon\}@korea.ac.kr).}
    \thanks{D. Kim is with the Institute of Advanced Technology Development (IATD), Hyundai Motor Company (e-mail: donghyeon.kim@hyundai.com)}
}
\maketitle

\begin{abstract}
Spurred by consistent advances and innovation in deep learning, object detection applications have become prevalent, particularly in autonomous driving that leverages various visual data. As convolutional neural networks (CNNs) are being optimized, the performances and computation speeds of object detection in autonomous driving have been significantly improved. However, due to the exponentially rapid growth in the complexity and scale of data used in object detection, there are limitations in terms of computation speeds while conducting object detection solely with classical computing. Motivated by this, quantum convolution-based object detection (QCOD) is proposed to adopt quantum computing to perform object detection at high speed. The QCOD utilizes our proposed fast quantum convolution that uploads input channel information and re-constructs output channels for achieving reduced computational complexity and thus improving performances. Lastly, the extensive experiments with KITTI autonomous driving object detection dataset verify that the proposed fast quantum convolution and QCOD are successfully operated in real object detection applications.
\end{abstract} 


\section{Introduction}\label{sec:1}

With the consistent advancement of deep learning, many deep learning-based applications have improved performance and become practical. These deep learning-based applications require significant computational power due to the expected increase in dataset sizes and algorithm complexity~\cite{DBLP:journals/access/ChenL14}. In particular, the computational complexity becomes more significant in object detection due to the growing complexity of data, which expands from 2D images to 3D point clouds and multi-modal data~\cite{DBLP:journals/pvldb/Whang020}. To cope with the growing complexity, several research aims to enhance the model architectures and the algorithmic advantages for improving the computation speed and performance of applications~\cite{DBLP:journals/tiv/KarimiKV23, DBLP:journals/tiv/AnsariND22}. In the era of classical computing, these algorithmic improvements yield highly positive results~\cite{DBLP:journals/tmm/ZhangCWL22,DBLP:journals/tmm/ChenDTCLG22}. However, they encounter fundamental challenges in efficiently conducting highly complex and complicated applications due to the inherent limitations of classical computing resources~\cite{DBLP:journals/tmm/WuLCHQ22}. A classical convolutional neural network (CNN) is one of the representatives that shows these limitations~\cite{DBLP:journals/tmm/RenK0PS21}. While CNN-based algorithms demonstrate rapid execution and viable performance, the computational complexity of CNN is significantly contingent upon the input size, with a computational cost of $\mathcal{O}(X \cdot C_{in} \cdot C_{out})$, where $X$, $C_{in}$, and $C_{out}$ denote the product of input data size and kernel size, input channel, and output channel, respectively~\cite{DBLP:journals/corr/OSheaN15}. This can pose a substantial challenge when applying the convolution process to extensive and intricate datasets, impeding its scalability in dealing with the rapidly expanding dataset. These challenges are exacerbated in complex applications based on CNNs, such as object detection. As the model's architecture grows in complexity and the dataset employed becomes more intricate, it is evident that relying solely on classical computing for such applications for real-time execution is not practical due to computational limitations~\cite{yan2018second}.

\begin{figure}[t]
    \centering
    \begin{tabular}{@{}c c@{}}
    \includegraphics[width=.32\columnwidth]{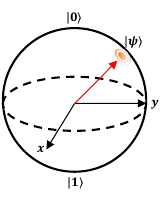}&
         \includegraphics[width=.49\columnwidth]{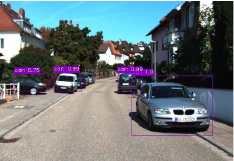}
         \\
         \small (a) 1-qubit system. & \small (b) Quantum object detection. \\
    \end{tabular}
    \caption{A brief illustration of quantum computing and its application.}
    \label{fig:1_q_od}
    \vspace{-3mm}
\end{figure}

Quantum computing is regarded as a promising solution to resolve these computational limitations. The emergence of the noisy intermediate-scale quantum (NISQ) era suggests that the number of available quantum bits, \textit{i.e.}, qubits, will exceed thousands by 2025, potentially achieving quantum advantage~\cite{arute2019quantum}. This achievement is based on the intrinsic nature of quantum computing, superposition, which enables quantum computing to excel in classical computing in various complex tasks. Fig.~\ref{fig:1_q_od} (a) represents an example of an 1-qubit superposition. In contrast to classical bit, the state of a qubit can be depicted as $\ket{\Phi} = \alpha_{0}\ket{0} + \alpha_{1}\ket{1}$, where $\alpha_{0}$ and $\alpha_{1}$ are the probabilistic complex amplitudes of qubit, satisfying $\alpha_{0}^2 + \alpha_{1}^2 = 1$. This representational capacity becomes increasingly more extensive as the number of qubits grows, because the number of bases, \textit{e.g.}, $\ket{00}$ and $\ket{01}$, expands to $2^q$, where $q$ denotes the number of qubits. On the other hand, there are challenges in implementing quantum computing-based applications in the NISQ era. As quantum computing and its algorithms are still in their early stages, there is a lack of optimization methods for tasks using quantum computing~\cite{DBLP:journals/corr/abs-2303-09491}. Additionally, quantum computing fails to replace classical computing entirely because it cannot perform structured tasks like convolution in CNNs~\cite{DBLP:journals/csur/AburaedKB17}. Moreover, the limited availability of deep learning techniques, datasets, and optimization tools for classical computing, along with the complexities of quantum computing, make the advancement of deep learning applications through quantum advantage in the realm of quantum machine learning challenging. 

Inspired by this, this paper focuses on the object detection, one of the most complicated applications using CNN. To cope with the growing complexity of the object detection and achieve faster operation time, this paper proposes quantum convolution-based object detection (QCOD). Fig.~\ref{fig:1_q_od} (b) shows a brief example result of QCOD. With our proposed quantum convolution, named fast quantum convolution, which boosts the encoding process. The fast quantum convolution optimizes the advantages of quantum computing. As our fast quantum convolution encodes multi-channel data into an identical quantum system, QCOD achieves fast quantum speed-ups. In addition, to leverage classical optimization schemes and deep learning techniques for QCOD, this paper proposes heterogeneous knowledge distillation, a modified version of knowledge distillation, to train the region proposal layer of QCOD. Knowledge distillation is a well-known training method that transfers the knowledge of the pre-trained teacher model to the un-trained student models. In this paper, heterogeneous knowledge distillation selects the pre-trained classical region proposal network and quantum convolution region proposal network as teacher and student model, respectively. Via heterogeneous knowledge distillation, QCOD addresses the lack of quantum optimization schemes in object detection.

Furthermore, this paper verifies the superiority of fast quantum convolution in QCOD and substantiates the probability of achieving quantum object detection in the near future through extensive experiments and ablation studies. It is difficult to definitively state that quantum object detection is superior to classical object detection in the view of performance. However, this paper observes that quantum object detection can be realized, and the proposed QCOD shows significant speed-ups in object detection. Serving as a foundational step toward the implementation of quantum object detection, this paper provides an outlook for future research in quantum object detection.

\BfPara{Contributions}
The major contributions of QCOD are as follows. First of all, This paper designs a novel quantum convolution named fast quantum convolution, considering the qubits' representation ability. The fast quantum convolution encodes multiple channels into quantum states and achieves quantum speed-ups. Second, this paper proposes heterogeneous knowledge distillation to leverage classical optimization schemes and the knowledge from classical pre-trained models, addressing the lack of knowledge in the quantum domain. Third, this paper verifies the superiority of the fast quantum convolution when utilizing with quantum random access memory (QRAM). Finally, this paper conducts numerous experiments and, to the best of our knowledge, implements the first quantum object detection.


\section{Related Work}\label{sec:Related}
This section introduces previous research that is closely aligned with our fast quantum convolution and QCOD development. The pivotal topics include i) quantum machine learning implementation, ii) data re-uploading, and iii) knowledge distillation for subset model training.

\BfPara{Quantum machine learning implementation}
The implementation of quantum machine learning hinges on two fundamental categories: i) optimizing qubits' representation abilities and ii) employing fast QRAM searching algorithms to minimize complexity. These properties enable quantum computing to outperform classical counterparts. Among these categories, the research related to our considering quantum convolutional neural network (QCNN) is as follows. \textit{Baek et al.}~\cite{baek2023stereoscopic} address the scalability limitation of available qubits in quantum convolution filters by incorporating these filters into massive 3D data classification applications. In this project, they leverage the concept of fidelity to achieve robust performance. \textit{Shen et al.}~\cite{DBLP:journals/corr/abs-2106-10421} focus on the architecture of classical CNN, replacing the fourier transform process of the CNN with a quantum circuit, thereby enhancing the speed of the entire CNN. Furthermore, several studies aim to realize quantum advantage by combining QCNN and QRAM. \textit{Oh et al.}~\cite{DBLP:conf/icoin/OhCKK21} implement QCNN on QRAM to store large-sized data. In addition, \textit{Kerenidis et al.}~\cite{DBLP:conf/iclr/KerenidisLP20} prove the quantum advantage when utilizing QCNN and QRAM with small errors.

\BfPara{Data re-uploading}
Data re-uploading is an encoding technique based on the quantum information theory that the states of qubits can represent multiple information~\cite{DBLP:journals/corr/abs-2003-01695}. \textit{P{\'{e}}rez{-}Salinas et al.}~\cite{DBLP:journals/quantum/Perez-SalinasCG20} firstly propose and prove the feasibility of data re-uploading using the 1-qubit system of quantum machine learning. \textit{Friedrich et al.}~\cite{DBLP:journals/corr/abs-2205-13418} combine data re-uploading techniques with QCNN to encode multiple data within a few qubits for avoiding barren plateaus~\footnote{The phenomenon of barren plateaus, a characteristic of quantum machine learning, impedes the trainability of quantum machine learning models~\cite{DBLP:journals/corr/abs-2010-15968}. Similar to the local minima in classical machine learning, barren plateaus give rise to problems where parameters are not efficiently optimized. In addition, it is well-known that the increase of the number of qubits induces barren plateaus~\cite{DBLP:journals/corr/abs-2011-10530}.}. \textit{Schuld et al.}~\cite{schuld2021effect} confirm that data re-uploading allows quantum models to represent progressively richer frequency spectra while using a limited number of qubits. In this paper, we propose channel uploading, a modified version of data re-uploading, to cope with numerous number of channels of practical object detection applications.

\BfPara{Knowledge distillation for subset model training} Knowledge distillation is a training method to handle variations in deep learning resources and enhance robust training in real-world applications~\cite{DBLP:conf/icpr/SarfrazAZ20}. Knowledge distillation is typically incorporated as a regularizer in the loss function, aiming to minimize the difference between the logits of the teacher model and those of the target student model. The target student model can conduct robust training by transferring pre-trained knowledge from the teacher to the student model.~\textit{Cui et al.}~\cite{DBLP:journals/tcsv/CuiWRCZ22} adopts a knowledge distillation regularizer as a loss function for semi-supervised learning, aiming to process real-world images. In this paper, we take a step further by employing knowledge distillation between models in a heterogeneous domain. We set a classical CNN-based model as the teacher model and our fast quantum convolution-based model as the student model.

\section{Quantum Machine Learning}\label{sec:primer}

Quantum machine learning is a machine learning framework that leverages the quantum advantages of quantum computing to address challenges previously tackled by classical neural networks. The quantum machine learning comprises encoding, parameterized quantum circuits, and decoding. In this section, we dive into each process within quantum machine learning essentials for constructing our proposed fast quantum convolution and QCOD.

\BfPara{Basic quantum operations}
In contrast to classical bits, which have deterministic values of either 0 or 1, quantum computing allows for the superposition of two states simultaneously. This unique characteristic is expressed using Dirac notation as $\ket{\Phi} \triangleq \sum^{2^{q}}_{k=1} \alpha_k \ket{k}$, where $\ket{k}$ represents a basis in the Hilbert space, and $\forall q \in \mathbb{N}[1, \infty)$ and $\sum^{2^{q}}_{k=1} |\alpha_k|^2 = 1$. The initialized q-qubit system can be expressed as $\ket{0}^{\otimes q}$. Similar to classical computer logic gates, quantum gates are operators capable of manipulating the state of qubits. From a physics perspective, operations of the quantum gates can be interpreted as transitions of qubit states on the Bloch sphere from one point to another~\cite{DBLP:conf/iclr/LandmanTDMK23}. Note that these quantum gates are in the form of unitary matrices. In this paper, we denote the unitary matrices used for encoding as $U_{E}$ and the unitary matrices used for training as $U_{T}$, depending on their purposes.

\BfPara{Encoding}
It is essential to convert classical information into quantum information to facilitate the integration of quantum machine learning with classical computing. This transformation is achieved through the implementation of the quantum gates denoted as $U_{E}$. Mathematically, $U_{E}$ can be expressed as a $2^q \times 2^q$ unitary matrix within a $q$-qubit quantum system. Therefore, with the classical data $\mathbf{x}$, the encoded $q$-qubit quantum states can be represented as $
    |\psi_\mathbf{x}\rangle 
    =  U_{\textit{E}} (\mathbf{x}) |0\rangle^{\otimes q}$, where the encdoed quantum state $\ket{\psi_{\mathbf{x}}}$ is on $2^q$-dimension Hilbert space. In this work, we design quantum gates $U_{E}$ without any trainable parameters to encode classical information into quantum states consistently.

\BfPara{Parameterized quantum circuit}
After encoding the classical information on the target quantum system, the parameterized quantum circuit (PQC) trains the parameter as in attention~\cite{DBLP:conf/nips/VaswaniSPUJGKP17} in classical machine learning. Each PQC has trainable unitary matrices $U_{T}$~\footnote{The general expression is a parameterized or variational unitary matrices. In this paper, we emphasize the significance of a PQC, which comprises trainable unitary matrices for a straightforward explanation.}, which comprises trainable rotation gates $R_\Gamma(\theta)$ and trainable \textit{Controlled-}$\Gamma$ gates $C\Gamma$, where $\forall \Gamma \in \{ X, Y, Z \}$. Here, $\theta$ denotes the trainable parameters where $\forall \theta \in [0,2\pi]^{|\theta|}$. Revisiting the encoded quantum states $\ket{\psi_{\mathbf{x}}}$, the output of the PQC can be expressed as $|\psi_\mathbf{x,\theta}\rangle = U_{\textit{T}}(\theta)|\psi_\mathbf{x}\rangle$. Note that  the trainable unitary matrices $U_{T}$ use trainable parameters and encoded quantum states as inputs.

\BfPara{Decoding}\label{sec:decode}
To use the transformed quantum state $\ket{\psi_{x,\theta}}$ with classical computing, we consider using the expectation quantum value~\cite{DBLP:journals/qip/Czerwinski22}. These expectation value can be designed as  $\langle O_{\mathbf{x},\theta}\rangle= \prod\nolimits_{M\in\mathcal{M}}\bra{\psi_{\mathbf{x},\theta}}M\ket{\psi_{\mathbf{x},\theta}}$, where $\langle O_{\mathbf{x}, \theta}\rangle$ denotes the expectation of quantum measured values on Hermitian matrices $M$. To decode the quantum information of each qubit, this paper designs $\mathcal{M}=\{M_l\}^q_{l=1}$, where $M_l = I^{\otimes l-1} \otimes Z \otimes I^{L-l}$. Here, $I$ denotes an identity matrix and $Z$ denotes a Pauli-Z matrix $ \small{\begin{bmatrix}
1 & 0 \\
0 & -1
\end{bmatrix}}$~\cite{DBLP:journals/corr/abs-2106-12627}. As a result, the output expectation values exist in $\langle O_{\mathbf{x},\theta}\rangle \in [-1,1]^{q}$.

\section{Fast Quantum Convolutional Neural Networks}\label{sec:fast quantum}
This section presents the details of our proposed fast quantum convolution, which can mitigate the computational overheads via patch processing, channel uploading, and channel reconstruction.

\begin{figure}[t]
    \centering
    \begin{tabular}{@{}c@{}c@{}c@{}c@{}}
    \includegraphics[width=.24\columnwidth]{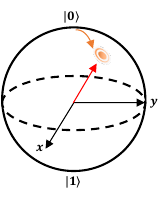}&
         \includegraphics[width=.24\columnwidth]{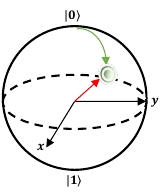} &
          \includegraphics[width=.24\columnwidth]{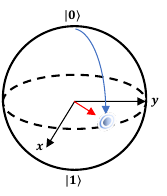}&
         \includegraphics[width=.24\columnwidth]{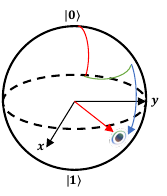}
         \\   
         \small (a) Red  & \small (b) Green  & \small (c) Blue  & \small (d) \textit{Proposed}\\
    \end{tabular}
     \vspace{-2mm}
    \caption{Comparison between existing channel uploading strategies (a-c) and our proposed channel uploading strategy (d).}
    \label{fig:channel_re-up}
\end{figure}

\begin{figure*}[t]
\centering
\includegraphics[width=.98\linewidth]{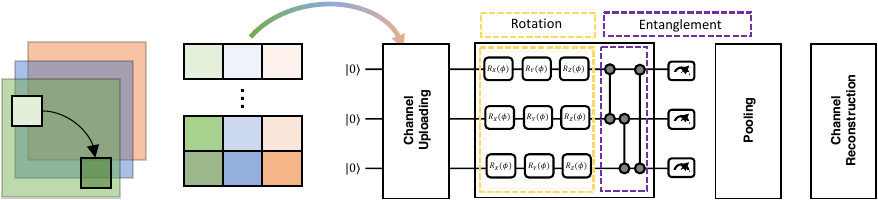}
\caption{An illustration of the proposed fast quantum convolutional neural network.}\label{fig:faster_quantum_convolution}
\end{figure*}

\begin{figure*}[t]
\centering
\begin{tabular}{p{.49\linewidth}p{.49\linewidth}}
   \includegraphics[width=\linewidth]{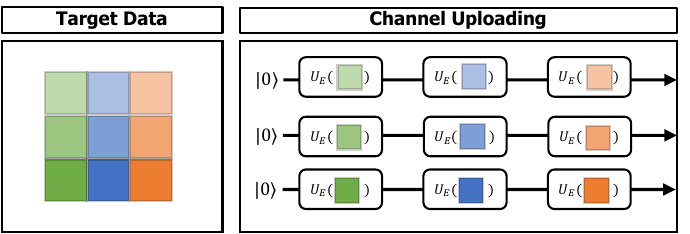} &  \includegraphics[width=\linewidth]{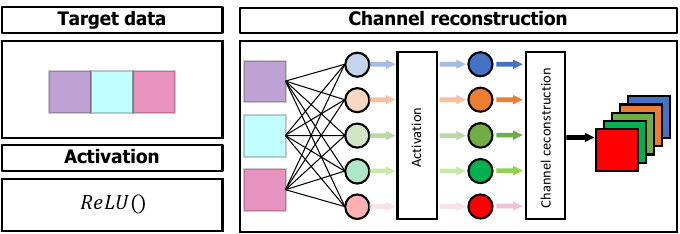} 
   \\
   \centering \small (a) The process of channel uploading. &  
   \centering \small (b) The process of channel re-construction. 
\end{tabular}
\caption{Detailed illustration of the proposed fast quantum convolution.}\label{fig:detail_conv}
\end{figure*}

\subsection{Motivation}

This paper aims to overcome the structural limitations of CNNs and existing QCNNs to cope with the growing complexity of real-world datasets in the field of object detection. To solve the limitations, we design our fast quantum convolution depicted in Fig.~\ref{fig:faster_quantum_convolution}. The fast quantum convolution employs i) patch processing, ii) channel uploading, iii) quantum feature extraction, and iv) channel reconstruction layer. Particularly, by employing channel uploading, we successfully mitigate the computational complexity. Fig.~\ref{fig:channel_re-up} briefly illustrates our proposed channel uploading strategy. In existing quantum computing systems, the approach of uploading data from the same channel is employed, as depicted in Fig.~\ref{fig:channel_re-up} (a-c). However, in real-world applications involving computations using numerous channels, a limitation arises where the operations must be repeated, corresponding to the number of channels ($C$). In contrast to the classical encoding scheme, our proposed channel uploading strategy focuses on the qubit's representation ability that is able to contain multiple pieces of information, depicted in Fig.~\ref{fig:channel_re-up} (d). Accordingly, our proposed fast quantum convolution can reduce entire computational complexity.

\subsection{Architecture of fast quantum convolutional neural network}
This paper employs three steps to design our fast quantum convolution: i) patch pre-processing, ii) quantum state encoding, and iii) quantum state decoding. The first step enables an optimized quantum convolutional process, while the other processes are necessary to attain quantum-induced image features. The overall process of fast quantum convolution is detailed in Algorithm~\ref{alg:fastq}.

\BfPara{Patch processing}
Inspired by classical image pre-processing optimization~\cite{DBLP:conf/ipps/RohwedderCAACW21}, this paper employs patch processing operation, named \textit{im2col}, which is widely implemented in classical CNN to fast quantum convolution. Fig.~\ref{fig:faster_quantum_convolution} illustrates the patch processing in our fast quantum convolution. In contrast to classical patch processing operations, our methods can reduce the number of operations by uploading the data in different channels in each qubit. A classical 3D tensor image $X^{l} \in \mathbb{R}^{H \times W \times C^l}$ is transformed to 2D tensor matrix $P^l \in \mathbb{R}^{(H^{l+1} W^{l+1})\times (HWC^{l})}$. Here, each patch in the input image is flattened to the part of each row of the 2D tensor matrix $P^l$. Therefore, the number of rows equals the one of operations.

\BfPara{Encoding via quantum channel uploading}
Based on the ability of qubits to maintain multiple pieces of information simultaneously, in contrast to classical bits, we sequentially upload the channels onto the same qubits. Fig.~\ref{fig:detail_conv} (a) provides a clear representation of quantum channel uploading. Each row of the 2D tensor matrix $P^l$ is uploaded on our quantum circuits. We use three different rotation gates $ R_x =\small{\begin{bmatrix}
\cos\left(\frac{\alpha}{2}\right) & -i\sin\left(\frac{\alpha}{2}\right) \\
-i\sin\left(\frac{\alpha}{2}\right) & \cos\left(\frac{\alpha}{2}\right)
\end{bmatrix}}$, $R_y  = \small{\begin{bmatrix}
\cos\left(\frac{\beta}{2}\right) & -\sin\left(\frac{\beta}{2}\right) \\
\sin\left(\frac{\beta}{2}\right) & \cos\left(\frac{\beta}{2}\right)
\end{bmatrix}}$ and $R_z  = \small{\begin{bmatrix}
e^{-i \frac{\delta}{2}} & 0 \\
0& e^{i \frac{\delta}{2}}
\end{bmatrix}}$. $\alpha$, $\beta$ and $\delta$ are constant values that can be modified according to the number of uploaded channels. Our proposed quantum channel uploading encoding strategy can be described as 
\begin{equation}
    |\psi\rangle_i = \prod_{j=0}^{HWC^l}\nolimits U_{E}(p^l_{i,j})|0\rangle^{\otimes q}, \label{eq:1}
\end{equation}
where $p^l_{i,j}$ denotes the components at the $i$-th row and $j$-th column of the 2D tensor matrix $P^l$. Note that $\forall i \in \mathbb{N}[0, H^{l+1}W^{l+1})$. Because the number of input components $HWC^l$ in each row is larger than the number of available qubits $q$ in the recent NISQ era, where the number of qubits is small, we design the encoding layer $U_{E}$ can employ additional data uploading strategy. Note that the quantum channel uploading process occurs in encoding, and the set of encoding layer $U_{E}$ doesn't have trainable parameters.

\BfPara{Quantum convolution} The fast quantum convolution utilizes the PQC to perform convolution on the information of the encoded quantum feature $\ket{\psi}_i$. This use of PQC can be compared with the convolution filters of classical CNNs. Unlike classical CNNs that employ element-wise products, PQC performs convolution through a trainable layer $U_{T}$ consisting of trainable \textit{Controlled-}$\Gamma$ gates and rotation gates $R_\Gamma(\theta)$, where $\forall \Gamma \in \{ X, Y, Z \}$. Particularly, by using trainable \textit{Controlled-}$\Gamma$, PQC is designed to induce mutual information referencing between qubits, creating entanglement~\cite{DBLP:journals/qic/XieL22}. This design allows the PQC to incorporate the spatial information of input data into the convolution more effectively. The quantum convolution can be represented as

\begin{equation}
    f(\ket{\psi}_i; \theta) : \ket{\psi_{\theta}}_i \leftarrow U_{T}(\theta)\ket{\psi}_i, \label{eq:2}
\end{equation}
where $\ket{\psi_{\theta}}_i$ is the output convoluted quantum states with trainable parameters $\theta$.

\BfPara{Quantum feature extraction via decoding}
Compared to classical CNNs, where convoluted features have discrete values, quantum features possess a probabilistic nature. This paper considers the quantum expectation value $\langle O_{\mathbf{x},\theta}\rangle$, represented in Sec.~\ref{sec:decode} as quantum features. To ensure stable learning in quantum computing for machine learning, we evaluate and utilize the probabilistic values associated with each basis' amplitude. In addition, as demonstrated in the example in Fig.~\ref{fig:1_q_od} (a), a quantum state can be represented as $\ket{\psi}_k = \alpha \ket{0} + \beta \ket{1}$, where $k \in \mathbb{N}[1, q]$, and it satisfies $\alpha^2 + \beta^2 = 1$. In this context, we set the output of quantum convolution as the probabilistic difference in amplitudes for each basis, i.e., $\alpha^2 - \beta^2$. This approach allows us to represent the output of each qubit as $\langle O_{\mathbf{x}, \theta} \rangle_{k} \in \mathbb{R}[-1, 1]$. We implement channel reconstruction layer to the entire output $\langle O_{\mathbf{x}, \theta} \rangle\in \mathbb{R}[-1, 1]^{\otimes q}$. Fig.~\ref{fig:detail_conv} (b) illustrates our channel reconstruction layers. By employing a linear function and an activation layer on the output, our fast quantum convolution succeeds in achieving scalability.

\begin{figure*}[t]
    \centering
    \includegraphics[width=1.2\columnwidth]{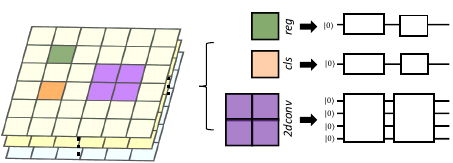}
       \caption{Proposed QRPN with our fast quantum convolution (The green and orange quantum convolution filter is designed for classification and box regression. The purple quantum convolution filter is designed for 2-dimensional convolution).}
    \label{fig:rpn}
\end{figure*}

\subsection{Strategy of fast quantum convolution}

\BfPara{Quantum backpropagation} Based on the quantum machine learning theory in~\cite{DBLP:journals/quantum/BanchiC21}, the gradients of quantum gates cannot be calculated directly. This is due to the intrinsic nature of quantum computing, where the output corresponds to the expectation of the corresponding computation outcome. Similarly, our fast quantum convolution outputs are also expressed as expectations, making the derivative of such expected outputs an invalid operation within the framework of quantum expectations. To solve this, this paper considers parameter-shift rule~\cite{DBLP:journals/quantum/WierichsIWL22}. 

\begin{equation}
    \frac{\partial \langle O_{\mathbf{x}, \theta}\rangle}{\partial {\theta}} = \frac{1}{2} [
        \langle O_{\mathbf{x}, \theta +\frac{\pi}{2}} \rangle - \langle O_{\mathbf{x}, \theta -\frac{\pi}{2}} \rangle 
        ],
\end{equation}
where $\langle O_{\mathbf{x}, \theta +\frac{\pi}{2}} \rangle$ and $\langle O_{\mathbf{x}, \theta -\frac{\pi}{2}} \rangle$ denote the extracted outputs with modified parameters $\theta +\frac{\pi}{2}$ and $\theta -\frac{\pi}{2}$, respectively.

\BfPara{QRAM for quantum speed-ups} As a counterpart to the random-access memory (RAM) in classical computing, quantum random-access memory (QRAM) stores the address of information in the state of the qubit. QRAM technology is a significant part to achieve quantum advantages and has drawn attention. One of the general QRAM structures is the bucket-brigade architecture, which inputs the address of the information in the quantum state and retrieves the data in the quantum state as the output~\cite{paler2020parallelizing}. The QRAM process of our considering fast quantum convolution can be described as

\begin{equation}
\sum_{i} |i\rangle_{address}|0\rangle_{data}\stackrel{\text{QRAM}}{\longrightarrow} \sum_{i} |i\rangle_{address}\left|\psi_{i}\right\rangle_{data},\label{eq:4}
\end{equation}
where $\ket{\cdot}_{address}$ and $\ket{\cdot}_{data}$ denote the storage address of QRAM and corresponding data, respectively. Based on the following Lemma~\ref{lemma:1}~\cite{DBLP:journals/corr/abs-2106-10421}, and Lemma~\ref{lemma:2}~\cite{DBLP:conf/iclr/KerenidisLP20}, we observe quantum speed-ups as depicted in Theorem~\ref{theorem:1}.

\begin{lemma}\label{lemma:1} \textbf{(Advantages of QRAM)}
Let target input $P \in \mathbb{R}^{n \times d}$, there exists a QRAM structure that conducts inserting, deleting, and updating each datum $p_{i,j}$ in time $\mathcal{O}(\log(n^2))$. In addition, there exists a quantum algorithm $ |i\rangle_{address}|0\rangle_{data} \rightarrow |i\rangle_{address}\left|\psi_{i}\right\rangle_{data}$ in time $\mathcal{O}(\log^2 n)$.

\end{lemma}

\begin{lemma}\label{lemma:2}
\textbf{(Running time of quantum gates)} 
Let symmetric matrix $M \in \mathbb{R}^{d \times d}$, datum $x \in \mathbb{R}^{d}$ and error $\delta >0$. When the matrix is stored in appropriate QRAM, there exists an algorithm that satisfies $\| \ket{z} - \ket{Mx} \|_{2} \leq \delta$ in time $\mathcal{O}((\sqrt{d}\kappa(M) + T_{x}\kappa(M))\log(1/\delta))$ with probability at least $1- \frac{1}{poly(d)}$, where $\kappa(M)$ and $T_x$ denote the condition number of $M$ and setting time for $\ket{x}$.

\end{lemma}

\begin{algorithm2e}[t]
\small
    \SetCustomAlgoRuledWidth{0.44\textwidth}  
\caption{Fast quantum convolution procedure.}
\label{alg:fastq} 
\textbf{Notation.} Input number of qubits: $q$, 3D tensor: $X^l$, transformed 2D tensor matrix: $P^l$, components at the $i$-th row, $j$-th column of the $P^l$: $p^l_{i.j}$, Channel Reconstruction function $T$ \;
\textbf{Input:} Input classical image $X^l$\;
\textbf{Patch processing.} $P^l \leftarrow X^l$\;

 \For{$i \in \{1, 2, \cdots, H^{l+1}W^{l+1} \}$}
    {   
        Initialize quantum state $\ket{0}^{\otimes q}$\;
        \For{$ j \in \{1, 2, \cdots, HWC^l  \}$}
        {
            $|\psi\rangle_i \leftarrow  U_{E}(p^l_{i,j})|0\rangle^{\otimes q}$\;
            $\ket{\psi_{\theta}}_i \leftarrow U_{T}(\theta)\ket{\psi}_i$\;
            \For{$k \in \{1,2, \cdots,  q \}$}
            {
                        Achieve $\langle O \rangle_{k} \in \mathbb{R}[-1, 1]$\;
            }
            
        }
                    
        Reshaping \& Pooling\;
            Channel Reconstruction $T : \mathbb{R}^{q} \rightarrow \mathbb{R}^{C^{l+1}}$\;              
                
    }
           
    Achieve extracted features $X^{l+1}$\; 
\textbf{Output:} Extracted features
\end{algorithm2e}

\begin{theorem}\label{theorem:1}
\textbf{(Running time of the fast quantum convolution)} Let the input classical 3D tensor image $X^{l} \in \mathbb{R}^{H^l \times W^l \times C^l}$ in $l$-th fast quantum convolution layer and the number of qubits $q = H^{l+1} \times W^{l+1}$ , where $\forall l \in L$ and $H^{l} \times W^{l} \geq 2$ and let the running time of encoding gates $U_{E}$ and $U_{T}$ as $T_{E}$ and $T_{T}$, respectively. With condition numbers of $l$-th encoding $\kappa(M_E)$ and PQC layer $\kappa(M_T)$ that satisfy $\kappa(M_{T}) = \rho^l \cdot  \kappa(M_{E})$, the running time of entire fast quantum convolution process $T_{total}$ is conducted in the time of
\begin{equation}
      \mathcal{O}(\log^2(H^{l+1}W^{l+1}) + \log(1/\delta)\kappa(M_E)(C^lt_E +\rho^lt_T)).
\end{equation}

\end{theorem}

\begin{proof}\label{proof2}
Based on the  patch processing, the 2D patch matrix can be generated as $P^l \in \mathbb{R}^{(H^{l+1} W^{l+1})\times (HWC^{l})}$. With \eqref{eq:4} and Lemma~\ref{lemma:1}, the total setting is conducted in time $T_{S} = \mathcal{O}(log(H^{l+1}W^{l+1})^2 + log^2(H^{l+1}W^{l+1})).$ Here, with the assumption $H^{l+1} \times W^{l+1} \geq 2$, then the time complexity of $T_S$ can be expressed as, 
\begin{equation}
    T_S \approx \mathcal{O}(log^2(H^{l+1}W^{l+1})).\label{eq:6}
\end{equation}
To conduct quantum convolution on the data in QRAM, it is necessary to design $M_{E} \in \mathbb{R}^{2^q \times 2^q}$ for encoding and $M_{T} \in \mathbb{R}^{2^q \times 2^q}$ for PQC. Based on Lemma~\ref{lemma:2}, the encoding process complexity and convolution process complexity can be described as
\begin{equation}
T_E = \mathcal{O}((\sqrt{2^q}\kappa(M_{E}) + C^{l} \cdot t_{E} \cdot \kappa(M_{E}))\log(1/\delta_E)).\label{eq:7}
\end{equation}
Similarly, the convolution process of fast quantum convolution can be described as 
\begin{equation}
    T_T = \mathcal{O}((\sqrt{2^q}\kappa(M_{T}) + t_{T} \cdot \kappa(M_{T}))\log(1/\delta_T)).\label{eq:8}
\end{equation}
Here, we observe that the trainable gate $t_T$ is called only once due to the advantages of channel uploading. For simplicity, by setting $\delta_E \approx \delta_T$ and $C^lt_E +\rho^lt_T \gg \sqrt{2^q}(1+\rho^l)$. The total time $T_{total} =T_S + T_E + T_T$ is conducted in time
\begin{equation}
     \mathcal{O}(\log^2(H^{l+1}W^{l+1}) + \log(1/\delta)\kappa(M_E)(C^lt_E +\rho^lt_T)).
\end{equation}

   On the other hand, with existing quantum convolution method, \eqref{eq:7}-\eqref{eq:8} is modified as  $T^{'}_E = \mathcal{O}(C^l(\sqrt{2^q}\kappa(M_{E}) +  t_{E} \cdot \kappa(M_{E}))\log(1/\delta_E))$ and
    $T^{'}_T = \mathcal{O}(C^l(\sqrt{2^q}\kappa(M_{T}) + t_{T} \cdot \kappa(M_{T}))\log(1/\delta_T))$, respectively. With a large number of channels $C^l$, we observe the advantage of proposed fast quantum convolution.
\end{proof}

\section{Quantum Object Detection}\label{sec:qod}
This section provides a detailed description of our method for implementing our fast quantum convolution in object detection. Note that the quantum version of the region proposal network (RPN) proposed below is a significant component of our QCOD.

\subsection{Motivation of quantum region proposal network}

The RPN is a major network utilized in object detection applications~\cite{DBLP:journals/corr/abs-1905-01614}. It is due to the role of RPN that localizes and proposes the target object. The RPN employs convolutional layers composed of spatial filters with dimensions $n \times n$, where $n \geq 1$. In addition, convolutional layers containing $1 \times 1$ spatial filters for box regression and classification are employed, respectively. Using these convolutional filters, the RPN takes an extracted feature as input, which is obtained from an extractor, and generates a set of rectangular object proposals, each of them accompanied by an objectness score. The outputs, \textit{i.e.}, object proposals and objectness scores, serve as the foundation for enabling the classifier to categorize objects within the target proposal. However, Despite excellent performance, RPN is still a computationally expensive network owing to the numerous number of proposed regions and channels involved in~\cite{DBLP:conf/icip/BappyR16}. 

\subsection{Architecture of quantum region proposal network}
To solve the limitations, this paper modifies the RPN~\cite{DBLP:journals/pami/RenHG017} to a quantum version of the RPN (QRPN) using our fast quantum convolution. QRPN aims to calculate object proposals and objectness scores using our fast quantum convolution. In contrast to RPN, which slides spatial filters on target tensor and utilizes element-wise multiplication, QRPN encodes multiple channel inputs jointly convolute the features via unitary gates (\textit{e.g.,}$R_x$, $R_y$ and $CNOT$ gates). By utilizing our fast quantum convolution, QRPN mitigates the time complexity as proved in Theorem~\ref{theorem:1}. In classical RPN structure, $1 \times 1$ convolution filters are utilized for computing two different loss functions (\textit{i.e.,} classification loss and box regression loss). Here, due to the $1 \times 1$ scale filter operation strategy, they can be considered as scalar-multiplied fully connected layers. To implement these characteristics of $1 \times 1$ filter-based convolution using fast quantum convolution, we design another structure of our fast quantum convolution. With a initialized single qubit $\ket{0}$ and each $1 \times 1$ quantum convolution can be expressed as 

\begin{equation}
     |\psi_{\theta}\rangle_i = \prod_{j=0}^{HWC^l}\nolimits  U_{1 \times 1}(\theta)\cdot U_E(p^l_{i,j} )|0\rangle^{\otimes 1}, \label{eq:1x1}
\end{equation}
where $\forall i \in \mathbb{N}{[0, H^{l}W^{l})}$, and $U_{1\times 1} \subset U_{T}$ denotes unitary matrices which is activated on each qubit. Note that the difference between $U_{T}$ and $U_{1\times 1}$ comes from the usage of 2-qubit gates, which can induce the entanglement in the designed quantum circuit.

\subsection{Heterogeneous knowledge distillation training}
Training the QRPN is challenging due to the limited availability of quantum computing resources and optimization tools, especially when considering object detection applications that rely on the pre-trained and well-optimized convolution-based RPN. Thus, to cope with these challenges and optimize QRPN-based object detection, we train QRPN using heterogeneous knowledge distillation. We set the pre-trained classical RPN and QRPN as a teacher and student model, respectively. Accordingly, well-optimized convolution knowledge of pre-trained model can be transferred to QRPN. Here, as the logits of fast quantum convolution are in range [-1, 1] and the logits of the classical convolution are in the range $(-\infty, \infty)$, we normalize both logits of the classical convolution and quantum convolution using \textit{ReLU}. Therefore, we make both logits are in the same space $[0, \infty)$. With the normalization, we design the classical to quantum (C2Q) loss function as
\begin{equation}
     \mathcal{L}_{\textit{C2Q}}(\thetab^Q) = \|\Omega(\mathbf{x};\thetab^Q) - 
    \Omega(\mathbf{x};\thetab^C)\|,
\end{equation}
where $\thetab^Q$ and $\thetab^C$ denote the trainable parameters of QRPN and classical RPN, respectively. $\Omega(\mathbf{x};\thetab^Q)$ and $\Omega(\mathbf{x};\thetab^C)$ denote the outputs of QRPN and RPN when the input tensor $\mathbf{x}$, respectively. Note that the dimensions and sizes of the inputs and outputs of QRPN are designed to be identical to those of RPN.

\subsection{Loss of QRPN}

As our QRPN is designed to be activated similarly to the classical RPN, which first regresses the box and then classifies the image within the box, we incorporate the heterogeneous knowledge distillation regularizer into each loss function of the QRPN ($L_{\text{cls}}$ and $L_{\text{reg}}$). The total loss functions are designed as 
\begin{equation}
  L_{\text{total}} = \frac{1-\gamma}{N_c}\sum_{c=1}^{N_c}L_{\text{cls}} + \frac{\lambda(1-\gamma)}{N_r}\sum_{r=1}^{N_r} L_{\text{reg}} + \gamma\mathcal{L}_{\textit{C2Q}}
\end{equation}
where $L_{\text{reg}}$ and $L_{\text{cls}}$ are formulated as presented in~\cite{DBLP:journals/pami/RenHG017}. In addition, $\gamma$ denotes the heterogeneous knowledge distillation parameter $0 \leq \gamma \leq 1$. Lastly, $\lambda$ denotes the normalization parameters between $L_{\text{reg}}$ and $L_{\text{cls}}$.

\subsection{Implementation details}

To realize and simulate our fast quantum convolution and its application, QCOD, we implement QCOD with the following details. Table.~\ref{tab:notation} shows our implementation details. Particularly, in all experiments, we set the number of channels $C = 64$ among 256 total channels utilized in~\cite{DBLP:journals/pami/RenHG017}. The \textit{2d-conv} illustrated in Fig.~\ref{fig:rpn} is designed as a 4-qubit PQC. Both \textit{cls} and \textit{reg} in Fig.~\ref{fig:rpn} are designed with 1-qubit PQC. To design PQC, this paper employs trainable \textit{U3CU3} layers which is composed of \textit{CNOT} gates and other rotation gates~\cite{DBLP:conf/hpca/WangDGLPCH22}. The trainable gate is employed for the 1-qubit PQC. As an activation layer, we utilize \textit{ReLU} function.

\begin{table}[t]
\centering
\scriptsize
    \caption{Notations and implementation details.}
    \label{tab:notation}
    \small
    \begin{tabular}{c|l}
        \toprule[1pt]
        \multicolumn{2}{c}{\textsf{Notations for quantum computing}}\\\midrule
        $|\psi_{x}\rangle$ & The quantum state encoded with data $x$. \\
         $q$ & The number of available qubits. \\
        $\langle O_{x} \rangle$ & The observable derived by quantum state $|\psi_{x}\rangle$. \\
         $\Gamma$ & Pauli-$\Gamma$ gate, \textit{e.g.}, $\Gamma \in \{X,Y,Z\}$. \\
        $R_\Gamma $ & Rotation-$\Gamma$ gate, \textit{e.g.}, $\Gamma \in \{X,Y,Z\}$. \\
        $C\Gamma $ & Controlled-$\Gamma$ gate, \textit{e.g.}, $\Gamma \in \{X,Y,Z\}$. \\
        $\mathcal{M}$ & Measurement operator. \\
        $I$ & Identity matrix. \\

        \midrule
        \multicolumn{2}{c}{\textsf{Notations for fast quantum convolution}}\\ \midrule
        $X^l$ & The classical 3D tensor image of $l$-th layer.\\
        $P^l$ & The transfomed 2D tensor matrix of $l$-th layer. \\
      
        $\ket{\psi}_i$ & The transformed $i$-th row quantum states. \\
        $U_{E}$ & The un-trainable encoding gates. \\
        $U_{T}$ & The trainable PQC gates.\\
        $U_{1\times 1}$ & The unitary matrices with 1-qubit.\\
          $(H^l,W^l,C^l)$ & (Height, width, channels) of $l$-th layer. \\
        
        \midrule
        \multicolumn{2}{c}{\textsf{Notations for QCOD.}}\\ \midrule
          $C$ & The number of activated channels $\{16, 32, \mathbf{64}\}$.\\
         $\gamma$ & The distillation coefficients $\{\mathbf{0.1}, 0.3, ..., 0.9 \}$. \\
         $\mathcal{L}_{(C2Q, cls, reg)}$ & The utilized losses (KD, cls, reg ). \\
        $\Omega(\mathbf{x};\thetab^Q)$ & Output logit from quantum convolution layers. \\
        $\Omega(\mathbf{x};\thetab^C)$ & Output logit from classical convolution layers.\\
      
        \bottomrule[1pt]
     \end{tabular}
\end{table}

\section{Performance Evaluation}\label{sec:exp}

First of all, this section introduces following three hypotheses those are the main items which shoould be verified and discussed. 
\begin{itemize}
    \item \BfPara{Hypothesis\,1} Our fast quantum convolution has the potential to incorporate multiple pieces of information.

    \item \BfPara{Hypothesis\,2} Channel uploading has advantages in reducing computational complexity rather than classical quantum information encoding strategy. 
    
    \item \BfPara{Hypothesis\,3} QCOD has trainability, and heterogeneous knowledge distillation has advantages in training QCOD. 
\end{itemize}
To corroborate Hypothesis 1, we visualize quantum states with corresponding encoded channels. As multi-qubit states are challenging to be described~\cite{DBLP:journals/corr/abs-2207-14135}, we utilize $1\times 1$ PQC. In addition, to compare the advantages of channel uploading by adjusting the number of channels uploaded to corroborate Hypothesis 2. Finally, to verify Hypothesis 3, we visualize the results of QCOD and measure the performance of QCOD with various training strategies.
\subsection{Setup}
This paper conducts all the experiments using classical computing with a Linux-based machine which has Intel i9-10990k, NVIDIA Titan X (2ea), and RAM 128GB. Python v3.8.10 and quantum computing simulation libraries (\textit{e.g.}, torchquantum v0.1.5 \cite{qoc2022wang}, pytorch v1.8.2 LTS) are employed for simulating the fast quantum convolution. For QCOD simulation, we use KITTI dataset~\cite{DBLP:journals/ijrr/GeigerLSU13} which is a well-known object detection dataset for autonomous driving. All of the comparative models are trained with 15 epochs. This paper utilizes a pre-trained feature extraction module and classification module using VGG-16~\cite{DBLP:journals/corr/SimonyanZ14a}. Each module is pre-trained using IMAGENET~\cite{DBLP:journals/corr/SimonyanZ14a}. The initial learning rate is set to $10^{-2}$.

\BfPara{Comparison techniques}
To corroborate the advantages of the fast quantum convolution and its application, QCOD, we designed various simulations involving classification and object detection. The brief explanation for each model used in simulations is as follows:

\begin{enumerate}
    \item \textit{FQC} \textit{(proposed)}: 
    \textit{FQC} \textit{(proposed)} is a simple convolution layer designed using the fast quantum convolution. The input features are encoded via channel uploading and the output features are decoded via channel reconstruction.
    \item \textit{QCOD (\textit{proposed})}: The \textit{QCOD (\textit{proposed})} is designed using the fast quantum convolution for object detection. This model is trained via heterogeneous knowledge distillation.
    \item \textit{QCOD} (\textit{w/o kd}):  \textit{QCOD (\textit{w/o kd})} is a ablation version of \textit{QCOD} 
 (\textit{proposed}). The heterogeneous knowledge distillation parameter $\gamma$ is set to $0$. 
    \item \textit{F-RCNN (baseline)}:
    \textit{F-RCNN (baseline)} is a baseline model to implement QCOD. F-RCNN (baseline) is also used as a teacher network to train QCOD (\textit{proposed}) via heterogeneous knowledge distillation.

    \item \textit{Quantum convolution}: Quantum convolution is designed as an existing quantum convolution with patch processing. This model utilizes PQC for each channel.
\end{enumerate}

\subsection{Evaluation Results}
\begin{figure}[t]
    \centering
    \begin{tabular}{c c}
    \includegraphics[width=.35\columnwidth]{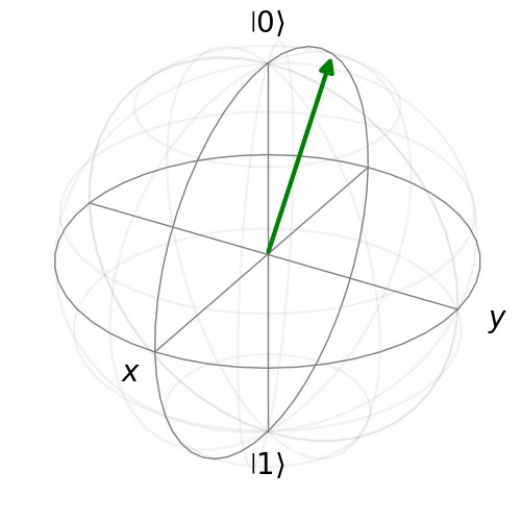}&
         \includegraphics[width=.35\columnwidth]{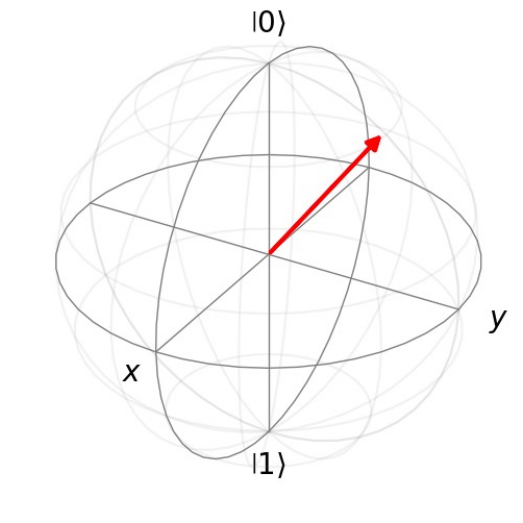} \\
          \small (a) Green ($R_x$).  & \small (b) Red ($R_y$).   \\
          \includegraphics[width=.35\columnwidth]{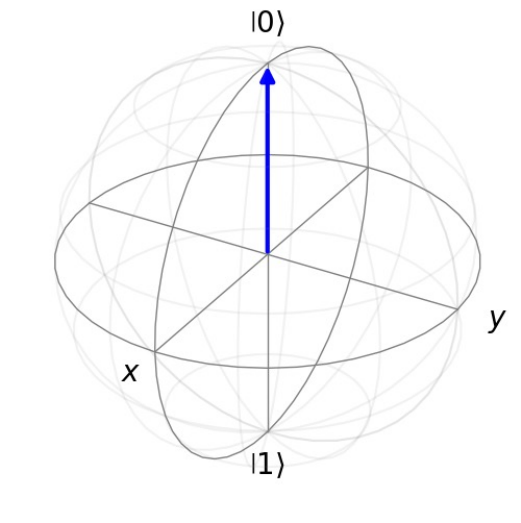} &
          \includegraphics[width=.35\columnwidth]{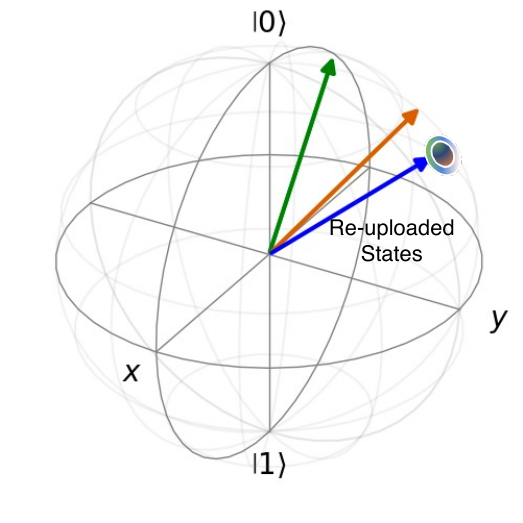} 
         \\   
         \small (c) Blue ($R_z$).  &\small (d) \textit{Proposed}. \\
    \end{tabular}
    \caption{Visualization of quantum states using various encoding strategies. (a-c) represents quantum states encoded with the classical quantum convolution strategy. (d) represents the states of fast quantum convolution. To visualize the quantum state, a $1$-qubit (PQC) is utilized. The (green, orange, blue) vectors in (d) represent the (first, second, final) quantum states with uploaded channels.}
    \label{exp:t-sne}
\end{figure}

\begin{figure}[t]
    \centering
    \includegraphics[width=.95\columnwidth]{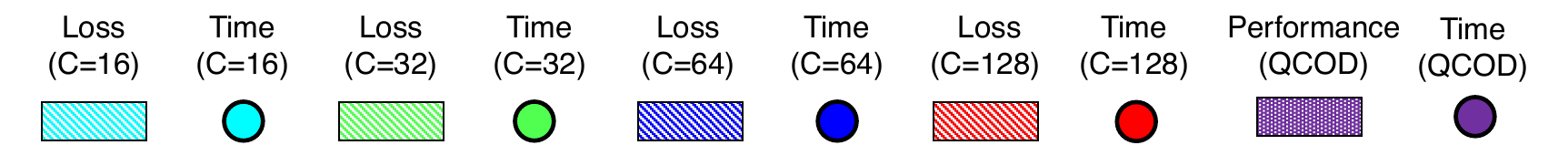}\\
    \includegraphics[width=.95\columnwidth]{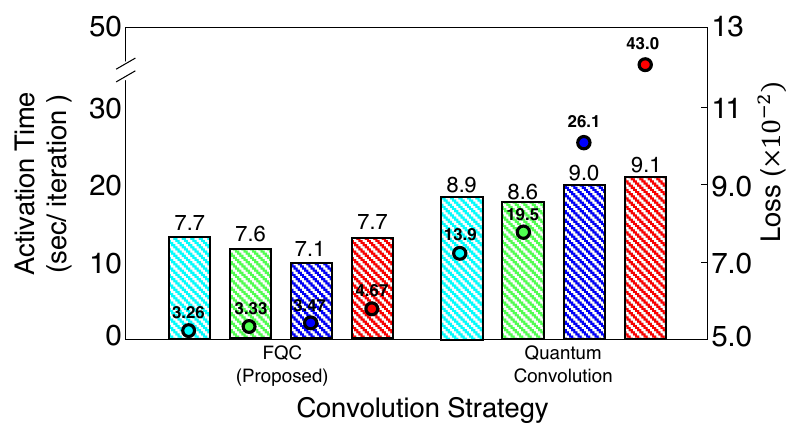}\\
    \small (a) Average iteration time and average loss of various convolution strategies.
    \begin{tabular}{c@{}c}
    \includegraphics[width=.47\columnwidth]{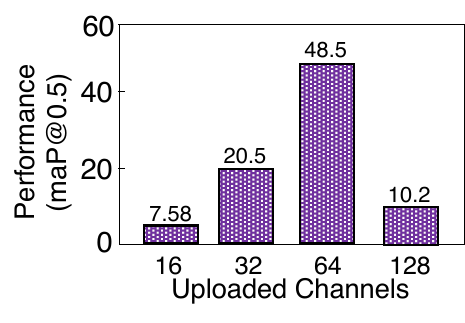} 
    &\includegraphics[width=.47\columnwidth]{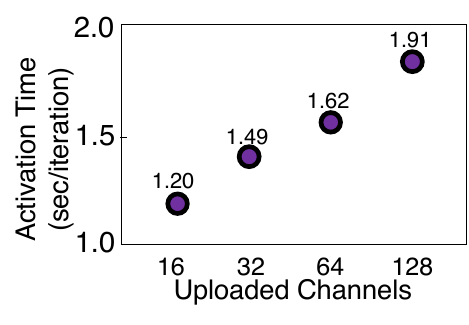} \\
    \small (b) Performance of QCOD.  & \small (c) Iteration time of QCOD.
    \end{tabular}
    \caption{Performance and loss comparison of various convolution strategies and their object detection applications (a) is conducted on the $32 \times 32$ size CIFAR10 dataset. (b) and (c) are measured on the resized $1382 \times 512 $ KITTI dataset with 1 batch-size. In experiments (b) and (c), we set knowledge distillation parameter $\gamma = 0.1$. We utilize the set of $R_y$ gates for uploading. }
    \label{exp:QCOD}
\end{figure}

\begin{figure*}[t!]
\centering
\begin{tabular}{p{.42\linewidth}p{.42\linewidth}}
   \includegraphics[width=\linewidth]{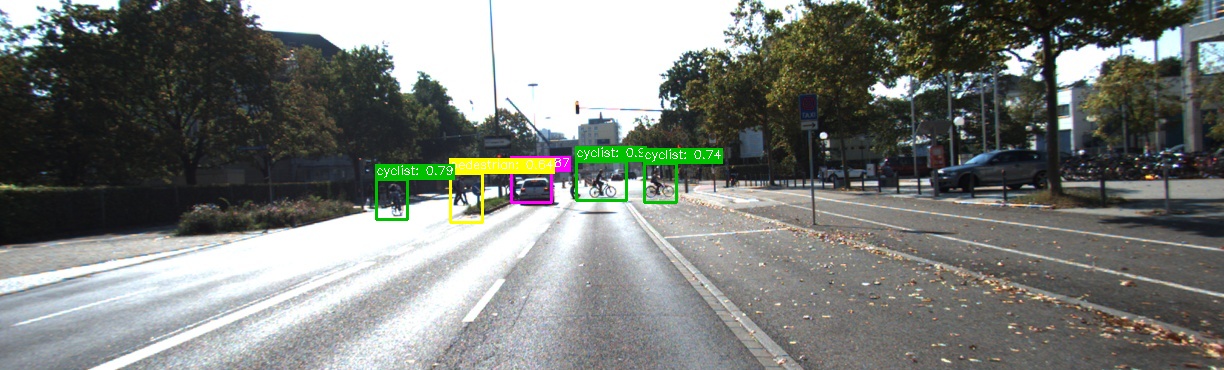} &  \includegraphics[width=\linewidth]{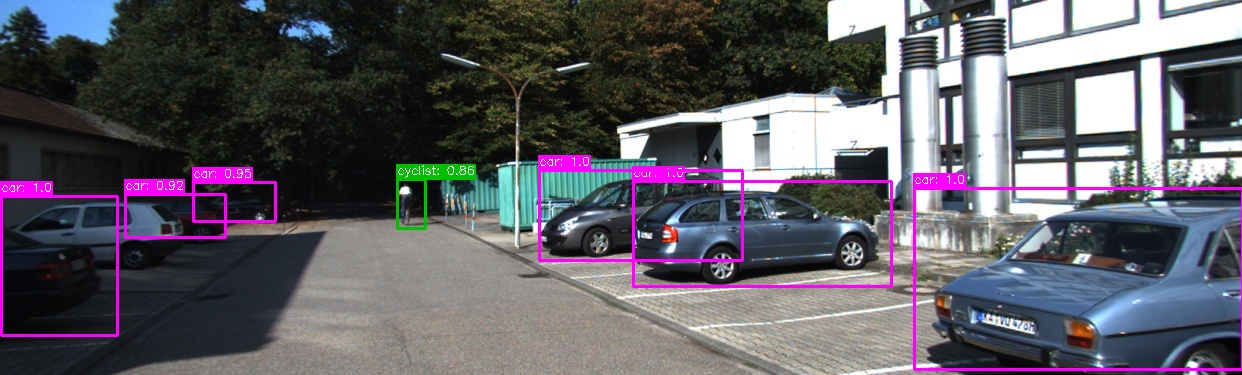}\\
   \includegraphics[width=\linewidth]{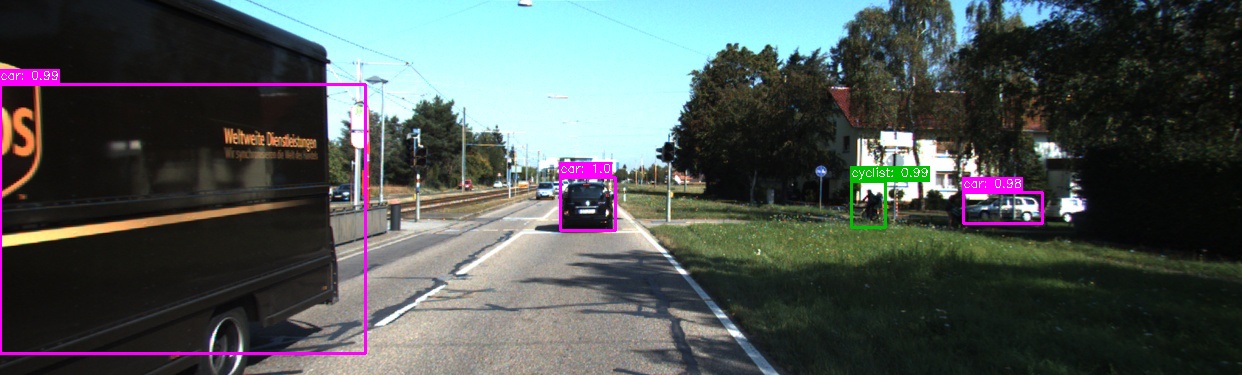} &  \includegraphics[width=\linewidth]{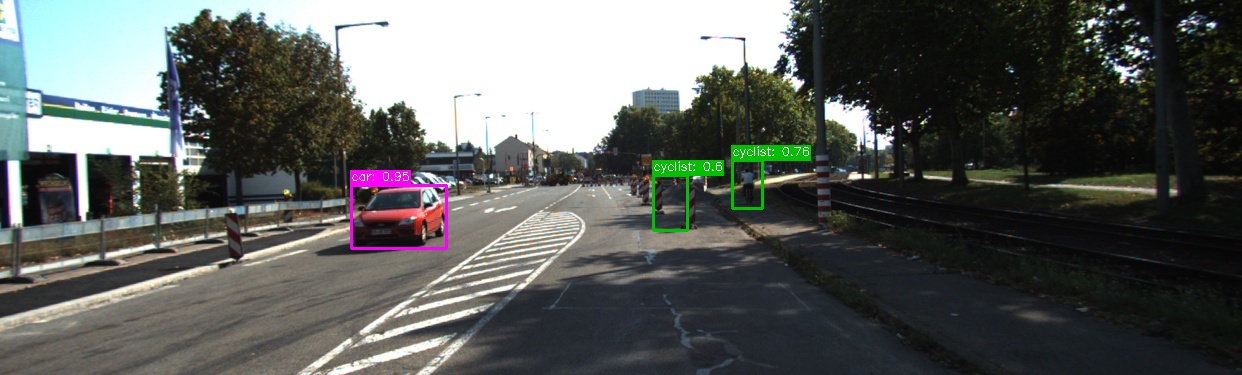} \\
\includegraphics[width=\linewidth]{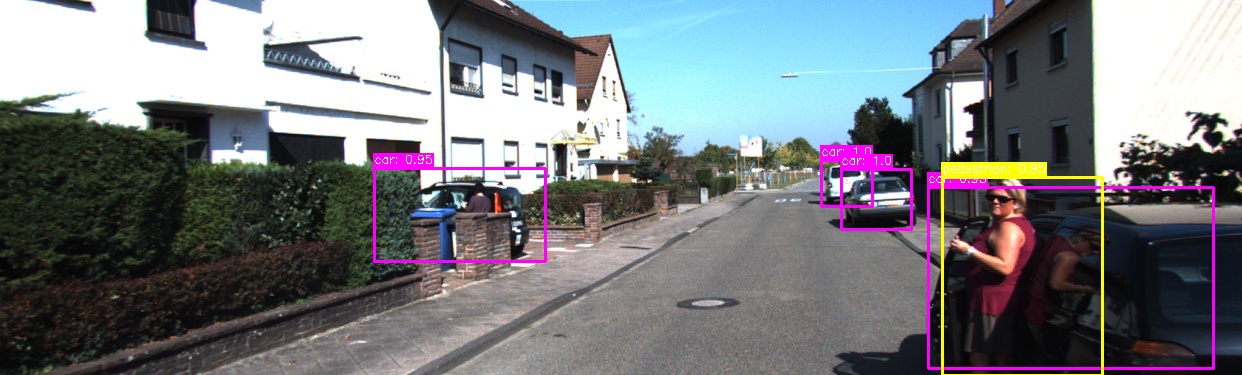} &  \includegraphics[width=\linewidth]{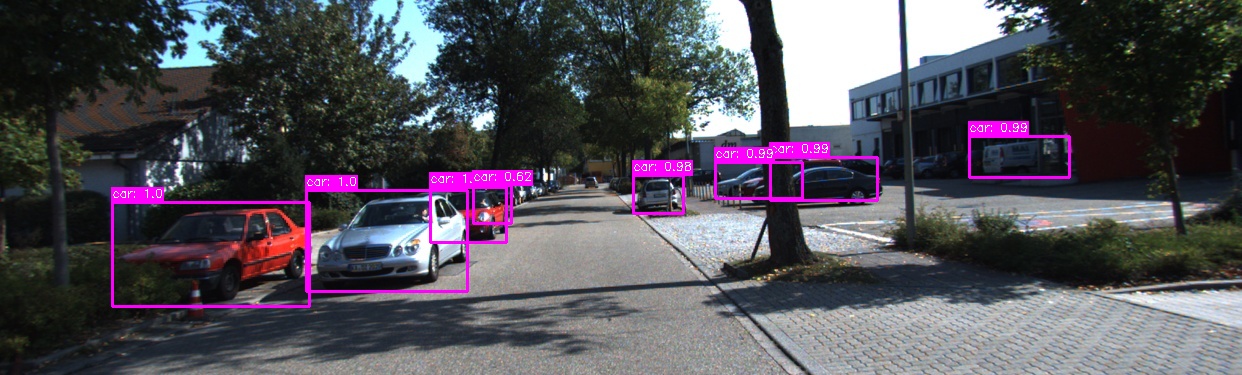} \\
 
   \includegraphics[width=\linewidth]{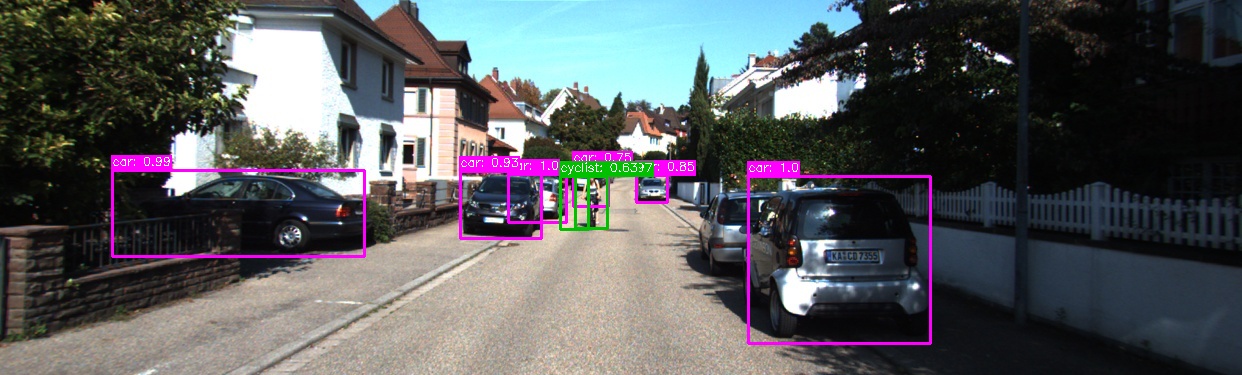} &  \includegraphics[width=\linewidth]{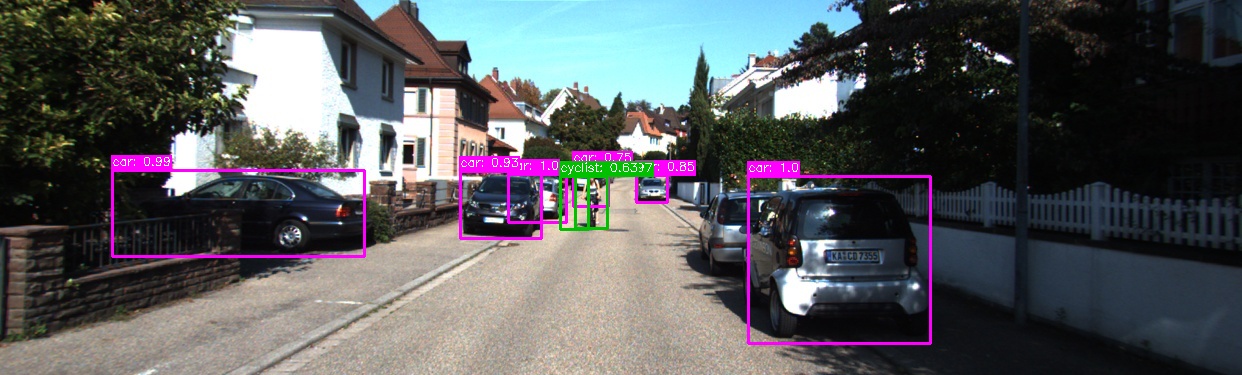} \\
   
   \includegraphics[width=\linewidth]{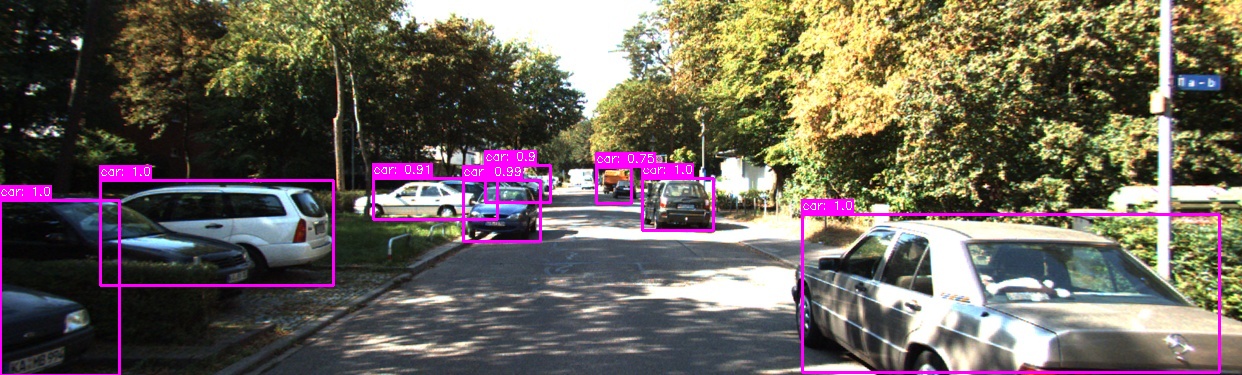} &  \includegraphics[width=\linewidth]{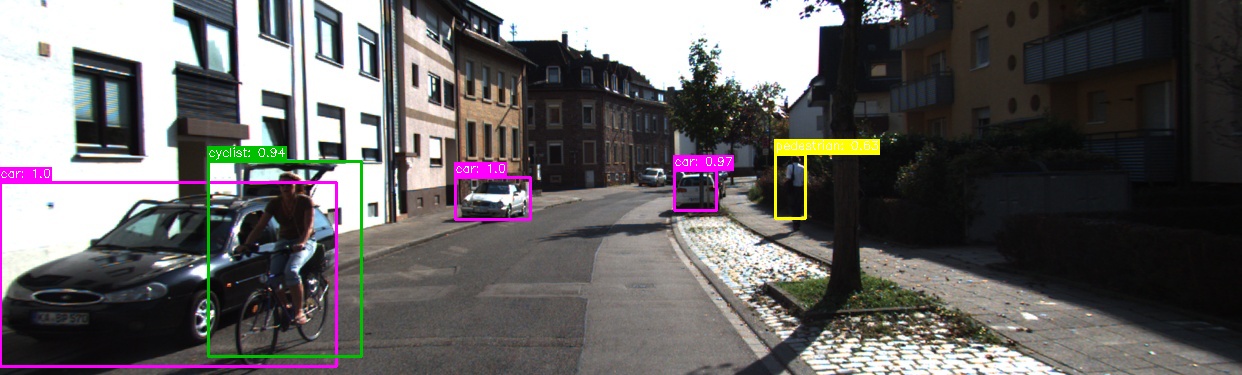} \\
   \includegraphics[width=\linewidth]{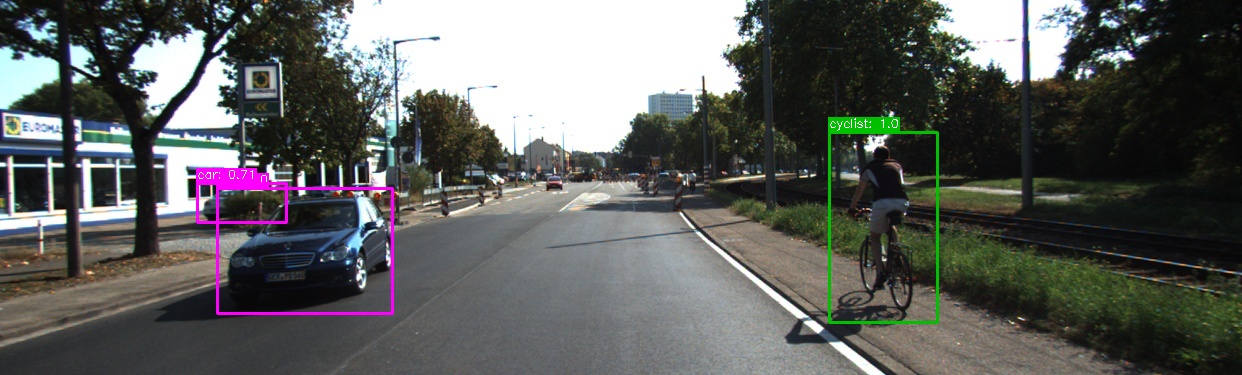} &  \includegraphics[width=\linewidth]{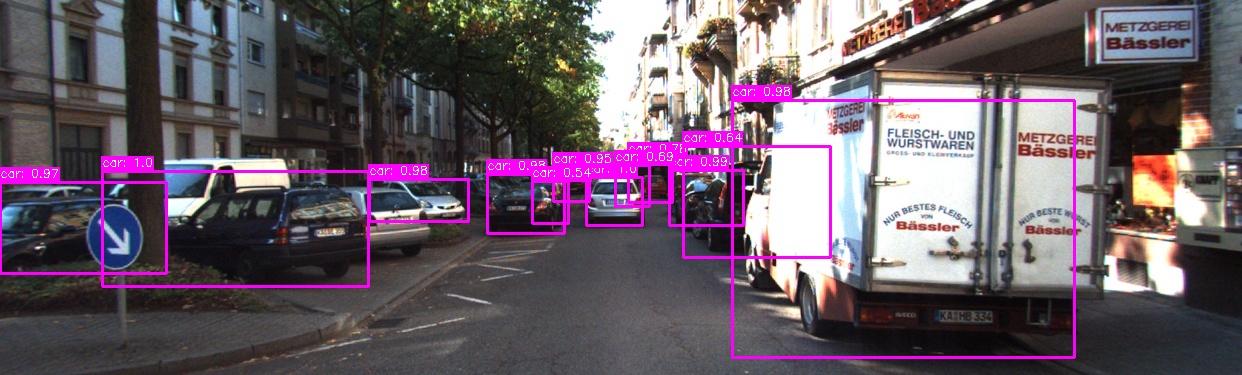} \\

   \includegraphics[width=\linewidth]{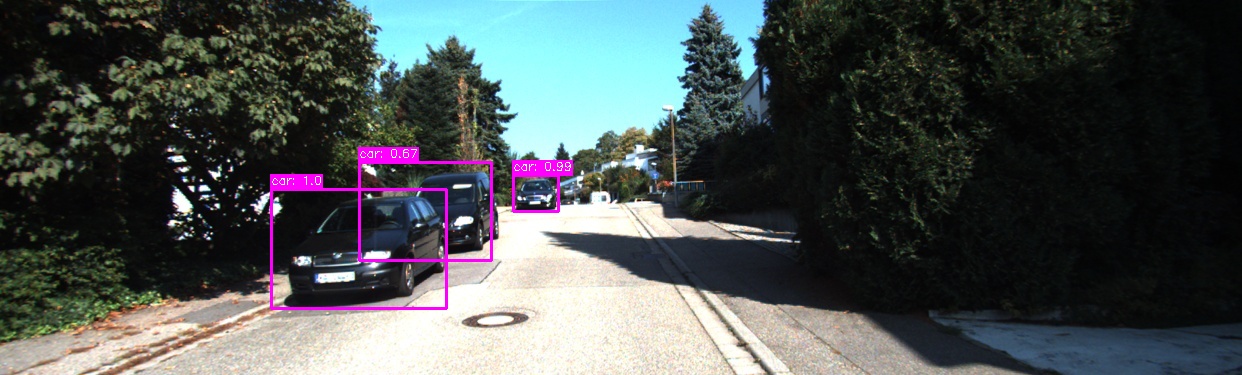} &  \includegraphics[width=\linewidth]{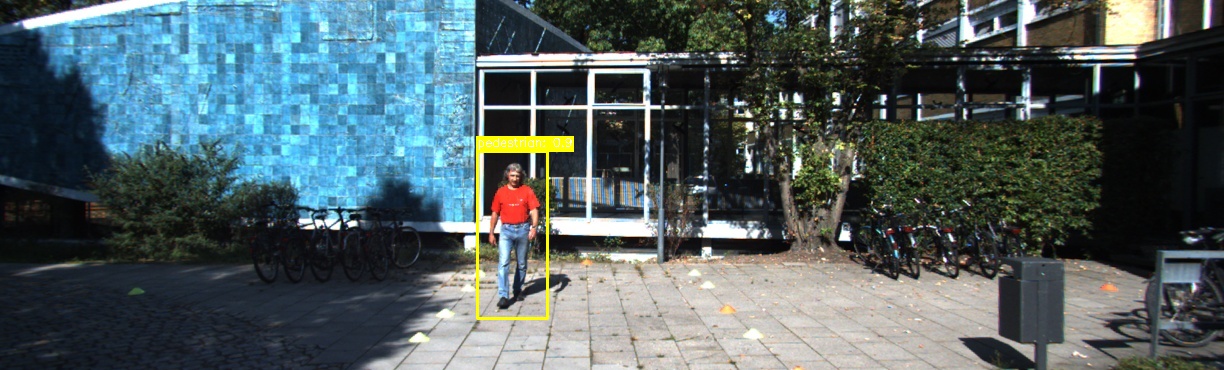} \\
   \includegraphics[width=\linewidth]{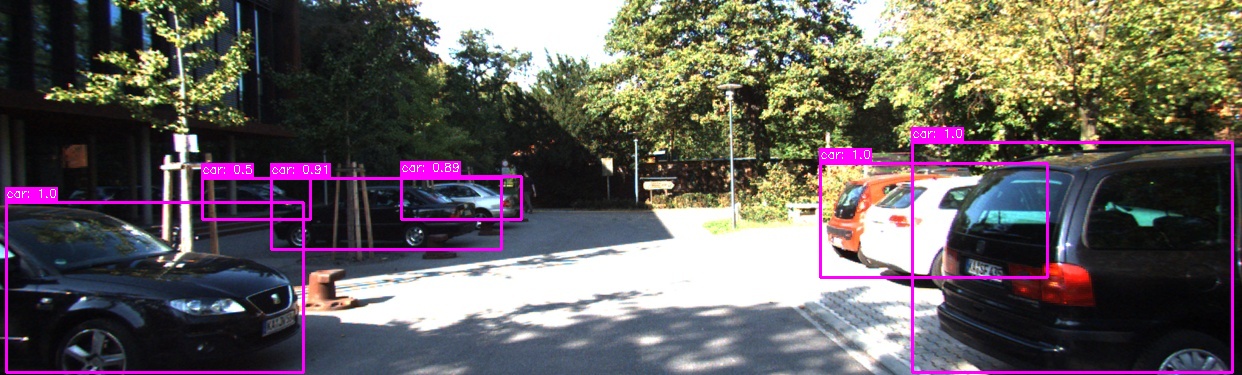} &  \includegraphics[width=\linewidth]{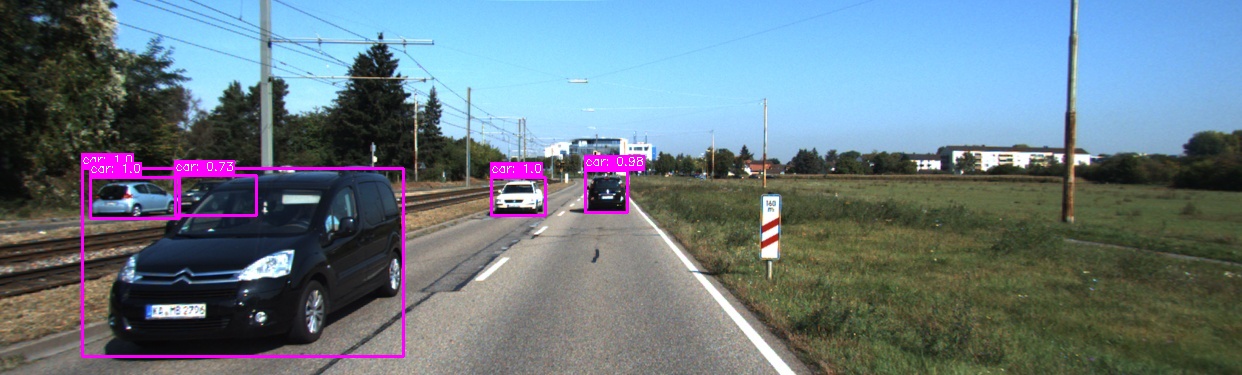} \\

\end{tabular}\caption{QCOD on KITTI dataset with heterogeneous knowledge distillation parameter $\gamma =0.3$ and with 64 number of channels. The yellow, green and pink boxes include pedestrians, cyclists and cars, respectively.}\label{exp:KITTI}
\end{figure*}

This paper corroborates the performance and feasibility of our fast quantum convolution through experiments represented in Fig.\ref{exp:t-sne} and Fig.\ref{exp:QCOD}(a). In addition, this paper conducts object detection computation using various training methods and numbers of channels in experiments, as represented in Table~\ref{tab:dataset} and Fig.\ref{exp:QCOD}(b-c). Finally, Fig.\ref{exp:KITTI} shows the results of object detection computation using our proposed QCOD with our fast quantum convolution on KITTI dataset. 

 \begin{table}[t]
\centering
\caption{MAP @0.5 (\%) of QCOD and comparative models\\ with 64 channels on KITTI dadaset. \\}
\vspace{-3mm}
\begin{tabular}{c|c||cccc}
    \toprule[1pt]  
     \multirow{2}{*}{ \textbf{Dataset}}& \multirow{2}{*}{ \textbf{Model}}&\multicolumn{4}{c}{\textbf{Heterogeneous KD ($\gamma$})}\\
    && $0$ & $0.1$  &  $0.3$ & $0.5$  \\\midrule
    \multirow{3}{*}{KITTI} &QCOD (\textit{proposed})& -& $48.5$& $\mathbf{51.2}$ & $49.1$ \\
    &QCOD (\textit{w/o KD})& $21.1$& -& - & -\\
     &F-RCNN (\textit{baseline})& $\mathbf{68.4 }$& -& -& -  \\
\bottomrule[1pt]
\end{tabular}\label{tab:dataset} 
\end{table}

\BfPara{Performance of fast quantum convolution} Fig.~\ref{exp:t-sne} and Fig.~\ref{exp:QCOD} (a) corroborate the feasibility and trainability of fast quantum convolution. We observe that features extracted from our fast quantum convolution in Fig.~\ref{exp:QCOD} (a) can be trained using classical optimization techniques. In Fig.~\ref{exp:t-sne} (a-c), the encoded states with classical quantum convolution achieve each channel information. Therefore, the number of required qubits equals the number of the input channels. However, As represented in Fig.~\ref{exp:t-sne} (d), our fast quantum convolution generates superpositioned quantum states that contain information about the channels. Furthermore, compared with Fig.~\ref{exp:t-sne} (c), our proposed model shows more clear representation ability even using same encoding gate. Based on the results presented in Fig.~\ref{exp:QCOD} (a), we observe that our fast quantum convolution executes more quickly than other quantum convolution methods. However, a significant decrease in performance is also observed as the number of uploaded channels exceeds the threshold value. This phenomenon is attributed to information loss resulting from the overlap of encoded information on qubits rather than a one-to-one encoding. It is a remaining challenge of our fast quantum convolution, and with the rapid advancement of quantum computing, this challenge can be mitigated as more qubits become available.

\BfPara{Feasibility of QCOD}
Through extensive experiments, we verify the feasibility of QCOD as an actual object detection application. Fig~\ref{exp:QCOD} (b) and (c) represent the performance and activation time of our QCOD according to the number of utilized channels. With the $64$ channels, our QCOD shows high performance, even in complex object detection application. As illustrated in Fig.\ref{exp:KITTI}, QCOD effectively draws bounding boxes tailored to objectives and transfers information, enabling object classification. As the first quantum version of object detection, QCOD shows the feasibility of quantum applications using our fast quantum convolution. To improve our QCOD, finding the optimal number of uploaded channels remains challenging. As QCOD increases the number of channels $32$ to $64$, the QCOD achieves $38\%$ performance gain. On the other hand, when the number of channels becomes $128$, a severe performance degradation is observed. In addition, Fig.~\ref{exp:QCOD} (c) represents that the performance and activation time are not proportional. Therefore, considering the performance and activation time, finding the optimal number of uploaded channels is crucial.

\BfPara{Advantages and disadvantages of heterogeneous knowledge distillation}
We investigate the impact of heterogeneous knowledge distillation on the performance of QCOD, which affects the practical utilization of quantum-based applications. Table~\ref{tab:dataset} shows the results of heterogeneous knowledge distillation. When we apply heterogeneous knowledge distillation with a parameter value of $\lambda = 0.3$, classical knowledge is effectively transferred, leading to an enhancement in QCOD performance. However, when we do not utilize heterogeneous knowledge distillation, our QCOD achieves a score of 21.1. These results highlight the substantial performance improvement can be achieved when utilizing heterogeneous distillation. On the other hand, when training with $\gamma > 0.3$, we observe a performance degradation. This result indicates a existence of the threshold value for $\gamma$, underscoring the significance of finding the appropriate $\gamma$ value tailored to the objectives of each QCOD applications.

\section{Concluding Remarks}\label{sec:conclusion}

This paper proposes the fast quantum convolution and its practical application in object detection, named QCOD. With our fast quantum convolution, which uploads input channel information and reconstructs output channel information, we observe the feasibility of quantum-based applications. To implement our fast quantum convolution in QCOD, we design a training method using heterogeneous knowledge distillation. By adopting knowledge distillation to transfer knowledge from the classical object detection domain to the quantum object detection domain, QCOD achieves robustness in object detection and adequately training the PQC. We analyze the complexity of our fast quantum convolution and QCOD to verify the advantages of our fast quantum convolution. Through extensive simulations, this paper corroborates i) the trainability of our fast quantum convolution, ii) the advantages of heterogeneous knowledge distillation, and iii) the feasibility of our QCOD.


\begin{IEEEbiography}[{\includegraphics[width=1in,height=1.25in,clip,keepaspectratio]{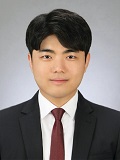}}]{Hankyul Baek} is currently a Ph.D. student in electrical and computer engineering at Korea University, Seoul, Republic of Korea, since March 2021, where he received his B.S. in electrical engineering in 2020. He was with LG Electronics, Seoul, Republic of Korea, from 2020 to 2021. He was also a visiting scholar at the Department of Electrical and Computer Engineering, The University of Utah, Salt Lake City, UT, USA, in 2023. 

His current research interests include quantum machine learning and its applications. 
\end{IEEEbiography}

\begin{IEEEbiographynophoto}{Dr. Donghyeon Kim} is currently a Principal Researcher at the Institute of Advanced Technology Development (IATD), Hyundai Motor and Kia Corporation, Republic of Korea. 
\end{IEEEbiographynophoto}

\begin{IEEEbiography}[{\includegraphics[width=1in,height=1.25in,clip,keepaspectratio]{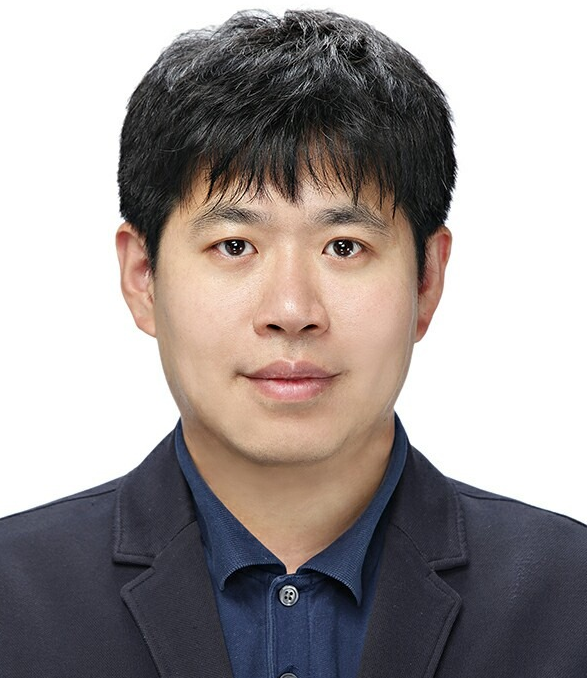}}]
{Prof. Joongheon Kim} (Senior Member, IEEE) has been with Korea University, Seoul, Korea, since 2019, where he is currently an associate professor at the Department of Electrical and Computer Engineering and also an adjunct professor at the Department of Communications Engineering (co-operated by Samsung Electronics) and the Department of Semiconductor Engineering (co-operated by SK Hynix). He received the B.S. and M.S. degrees in computer science and engineering from Korea University, Seoul, Korea, in 2004 and 2006; and the Ph.D. degree in computer science from the University of Southern California (USC), Los Angeles, CA, USA, in 2014. Before joining Korea University, he was a research engineer with LG Electronics (Seoul, Korea, 2006--2009), a systems engineer with Intel Corporation Headquarter (Santa Clara in Silicon Valley, CA, USA, 2013--2016), and an assistant professor of computer science and engineering with Chung-Ang University (Seoul, Korea, 2016--2019). 

He serves as an editor and guest editor for \textsc{IEEE Transactions on Vehicular Technology}, \textsc{IEEE Internet of Things Journal}, \textsc{IEEE Transactions on Machine Learning in Communications and Networking}, \textsc{IEEE Communications Standards Magazine}, \textit{Computer Networks}, and \textit{ICT Express}. He is also a distinguished lecturer for \textit{IEEE Communications Society (ComSoc)} and \textit{IEEE Systems Council}. He is an executive director of the Korea Institute of Communication and Information Sciences (KICS). He was a recipient of Annenberg Graduate Fellowship with his Ph.D. admission from USC (2009), Intel Corporation Next Generation and Standards (NGS) Division Recognition Award (2015), 
\textsc{IEEE Systems Journal} Best Paper Award (2020), IEEE ComSoc Multimedia Communications Technical Committee (MMTC) Outstanding Young Researcher Award (2020), IEEE ComSoc MMTC Best Journal Paper Award (2021), Best Special Issue Guest Editor Award by \textit{ICT Express} (2022), and Best Editor Award by \textit{ICT Express} (2023). He also received several awards from IEEE conferences including IEEE ICOIN Best Paper Award (2021), IEEE Vehicular Technology Society (VTS) Seoul Chapter Awards (2019, 2021, and 2022), and IEEE ICTC Best Paper Award (2022). 
\end{IEEEbiography}
\end{document}